\documentclass[journal,onecolumn,draftclsnofoot,12pt]{IEEEtran}

\newcommand{\etal}{et al.~}
\newcommand{\rnote}[1]{\textcolor{red}{\textbf{#1}}}
\newcommand{\bee}[1]{\ensuremath{\mathbb{B}^{#1}}}
\newcommand{\rf}[1]{(\ref{#1})}

\usepackage[utf8]{inputenc}
\usepackage[T1]{fontenc}
\usepackage{amsmath,amssymb,amsfonts}
\usepackage{algorithmic}
\usepackage{algorithm}
\usepackage{graphicx} 
\usepackage{textcomp}
\usepackage[outerbars]{changebar}
\usepackage[english]{babel}
\usepackage[dvipsnames,svgnames,x11names]{xcolor}
\usepackage[marginparwidth=2.5cm]{geometry}
\usepackage[markup=underlined]{changes}
\usepackage[hyphens,spaces]{url}
\usepackage{float}
\usepackage[caption = false]{subfig}
\usepackage{makecell}
\usepackage{setspace} 

\begin{document}

\title{Optimizing Occupancy Sensor Placement in Smart Environments}

\author{Hao Lu and Richard J.~Radke}


\maketitle

\markboth{IEEE Journal}{ temp }


\begin{abstract}

Understanding the locations of occupants in a commercial built environment is critical for realizing energy savings by delivering lighting, heating, and cooling only where it is needed. The key to achieving this goal is being able to recognize zone occupancy in real time, without impeding occupants' activities or compromising privacy. While low-resolution, privacy-preserving time-of-flight (ToF) sensor networks have demonstrated good performance in zone counting, the performance depends on careful sensor placement. To address this issue, we propose an automatic sensor placement method that determines optimal sensor layouts for a given number of sensors, and can predict the counting accuracy of such a layout. In particular, given the geometric constraints of an office environment, we simulate a large number of occupant trajectories. We then formulate the sensor placement problem as an integer linear programming
(ILP) problem and solve it with the branch and bound method.  We demonstrate the effectiveness of the proposed method based on simulations of several different office environments.

\end{abstract}


\section{Introduction}

Building energy consumption has become the main source of urban energy usage \cite{crawley2008contrasting}. The major contributors to energy consumption in commercial buildings are lighting and Heating, Ventilating, and Air-Conditioning ({HVAC}). Finding methods to minimize these costs will not only create significant energy savings for businesses, but also reduce emissions attributed to power generation. However, thermal comfort and air quality should not be neglected. Many studies \cite{sun_review_2020, odwyer_smart_2019} show that understanding where and when spaces are occupied is the key to balancing thermal comfort and energy savings. 

Deploying sensor networks to monitor occupancy information in buildings is a common approach.  For example, in our previous work \cite{lu2021zone} we demonstrated that a low-cost, low-resolution array of time-of-flight (ToF) sensors could accurately count occupants in typical office spaces. However, the counting accuracy of this, or any, sensor system depends not only on the counting algorithm, but also on the sensor placement layout. In our initial work, we manually determined the locations of the sensors based on intuition and trial and error.  Clearly this is suboptimal since the overall system accuracy should not depend on the expertise of the installer.

In this paper, we propose an automatic sensor placement method that determines optimal sensor layouts for a given number of sensors, and can predict the counting accuracy of such a layout. This helps designers estimate installation costs and expected counting performance prior to installation.  The algorithm only requires basic prior knowledge of the installation environment in the form of an annotated floor plan (e.g., walls, major furniture, and HVAC zone boundaries) to generate the optimal sensor placement.  The optimization algorithm is based on a simulation of occupant trajectories, with the objective of placing sensors to maximally cover occupant paths near zone boundaries.  The major contributions of this paper are as follows:
\begin{enumerate}
    \item We propose a method to simulate many realistic occupant trajectories based on an annotated floor plan of an office space. 
    \item We formulate the sensor placement problem as an integer linear programming (ILP) problem and solve it with the branch and bound method.
    \item We evaluate the sensor placement using the Unity game engine, coupled with a digital twin of the sensing system, comparing predicted to actual performance in several scenarios.
\end{enumerate}

The paper is structured as follows. We review recent research about optimizing sensor placement in different scenarios in Section \ref{sec:related}. In Section \ref{sec:traj}, we describe how occupant trajectories are created from the annotated floor plan, which provide the information necessary to formulate the optimization problem in Section \ref{sec:opt}.  We evaluate the overall system in Section \ref{sec:results}, and conclude in Section \ref{sec:conclusions} with discussion and ideas for future work.

\section{Related Work}
\label{sec:related}
The problem of Optimizing Sensor Placement (OSP) has been widely studied in diverse areas including wireless communications \cite{guo_energy-efficient_2020,karimi-bidhendi_energy-efficient_2021}, structural health monitoring systems \cite{yi_methodology_2012,shuo_3d_2016}, building environment monitoring \cite{yoganathan_optimal_2018}, indoor occupancy detection \cite{vlasenko_smart-condo_2015,wu_sensor_2020}, freeway traffic \cite{danczyk_mixed-integer_2011,danczyk_probabilistic_2016}, and water distribution systems \cite{hu_survey_2018}.  Depending on the scenario, OSP objective functions involve some combination of installation/energy cost, communication capacity, coverage of the sensor system, and detection accuracy. 

One of the most common OSP application areas involves wireless sensor networks (WSN). Research in this area is generally focused on reducing energy cost without compromising communication capacity, since most wireless sensors are powered by batteries that cannot be easily replaced. Guo \etal\cite{guo_energy-efficient_2020} proposed a routing-aware Lloyd-like algorithm for the sensor deployment problem in a homogeneous WSN to minimize energy cost. Karimi-Bidhendi \etal\cite{karimi-bidhendi_energy-efficient_2021} further discussed the convergence of this type of algorithm in a heterogeneous two-tier WSN whose optimal cells are non-convex.

In other scenarios, researchers are more focused on finding the smallest number of sensors required to guarantee a certain level of detection accuracy. For example, Yi \etal\cite{yi_methodology_2012} proposed a method to reduce the number of sensors required for the health monitoring of civil infrastructure using an ant colony algorithm. Shuo \etal\cite{shuo_3d_2016} further proposed a 3D sensor placement strategy to reduce the number of sensors required to monitor a bridge's structural health. Yoganathan \etal\cite{yoganathan_optimal_2018} studied the sensor placement problem for monitoring office building environments. They first densely placed sensors in the environment and then used clustering algorithms to remove the redundant sensors. Danczyk \etal\cite{danczyk_mixed-integer_2011, danczyk_probabilistic_2016} studied the optimal allocation of limited sensor resources to optimize travel time estimation in a freeway. They formulated the problem as a mixed-integer linear program and used a branch-and-bound tree to solve it. Wang \etal\cite{wang_reinforcement_2020} proposed a general optimization framework for the sensor placement problem using reinforcement learning.

In this paper, our focus is on indoor occupancy detection. While many studies address the trade-off between the number of sensors and detection accuracy, the variety of sensor types and detection requirements make the modeling and solution methods of the OSP problem for indoor occupancy detection very diverse. Wu \etal\cite{wu_sensor_2020} studied the indoor localization problem using the received signal strength indicator (RSSI) in wireless networks. They optimized the sensor placement to maximize the coverage using a genetic algorithm. Nguyen \etal\cite{nguyen_efficient_2020} proposed a greedy algorithm for deploying sensors for spatio-temporal environmental monitoring, which can minimize the prediction uncertainty. Fanti \etal\cite{fanti_smart_2017,fanti_integrated_2018} studied the OSP for  occupant location tracking using PIR sensors in the home environment. They formulated the problem as an integer linear programming (ILP) problem and discussed different objective functions to minimize the sensor number and maximize the detection accuracy. Gungor \etal\cite{gungor_respire_2021} further discussed the robustness of an entire system where sensors may break down or malfunction. They proposed a probabilistic model and also formulated it as an ILP. Vlasenko \etal\cite{vlasenko_smart-condo_2015} proposed a single person localization method using PIR sensors. They proposed a greedy method to optimize the sensor placement to improve the localization accuracy. Notably, this work assumes the occupants travel along typical path segments between rooms and that these path segments can be learned once given the floor plan. This assumption is consistent with our observations and is widely discussed in the field of human trajectory prediction \cite{alahi2016social,lee2022muse}. The way that Vlasenko \etal modeled typical path segments inspired us to generate occupant trajectories to pose and solve the OSP problem in our setting.  We also approach the formulation of the OSP as an ILP, similar to \cite{fanti_smart_2017, fanti_integrated_2018, gungor_respire_2021}.

\section{Occupant Trajectory Modelling}
\label{sec:traj}
The first step of our proposed algorithm is the generation of a large number of plausible occupant trajectories from an annotated floor plan.  We also require foreknowledge of the HVAC zone boundaries (which are likely known at design time based on the positions of independently controllable fans and vents), as well as a simple model for the sensor system to be installed (in particular, the view angle of each sensor).  None of this information should be particularly time-consuming to collect or require specific expertise.

We first assume that occupants' daily routines can be well-estimated from the floor plan of the office, based on the observation that occupant motion typically consists of several short paths between certain pairs of interest points depicted in the floor plan (e.g., office to restroom or entrance door to refrigerator).  We also assume that the sensor system is sufficiently accurate to estimate the location and velocity of an occupant if they pass through the field of view of a sensor.

Our overall approach is to (1) process a floor plan annotated with areas of interest into a grid of possible sensor locations, (2) estimate a large set of occupant trajectories based on pairs of interest points, and (3) formulate an optimal sensor placement problem based on selecting grid locations for sensors that cover the zone boundaries where occupants are likely to walk.

\subsection{Floor Plan Labeling}\label{section3.2}

\begin{figure}[htbp!]
\centering
\begin{tabular}{c}
\includegraphics[width=.9\linewidth]{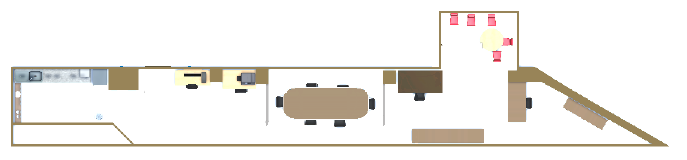}\\
(a)\\
\includegraphics[width=.9\linewidth]{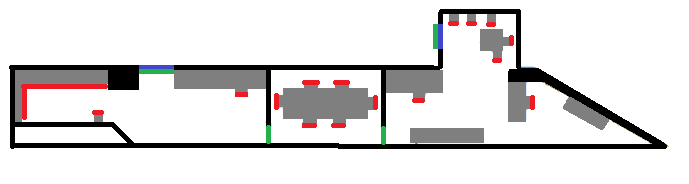} \\
(b) 
\end{tabular}
\caption{(a) An example floor plan of an office suite. (b) The color-labeled floor plan of the office suite. There are 6 labels: walls (black), obstacles (grey), doorways (brown), areas of interest (red), zone boundaries (green), and walkable regions (white).}
\label{fig:f1_fp}
\end{figure}

Inspired by Vlasenko \etal\cite{vlasenko_smart-condo_2015}, we first require the user to annotate the floor plan with 6 labels: walkable regions, walls, obstacles, doorways, areas of interest, and zone boundaries.  This markup can be quickly accomplished using a simple paint program, as illustrated in Figure \ref{fig:f1_fp}.  \rnote{A different color than brown would be better for the doorways (e.g., light blue).}

Walls and obstacles are not walkable for occupants, and can be directly obtained from the floor plan. Areas of interest are the locations where  occupants are likely to begin or end their movements. These areas include seats, restrooms, refrigerators, windows, switches, cabinets, and so on; annotating these regions is the main effort required of the user.  

Once we label the floor plan, we convert it into a uniform grid with a user-specified step size. In subsequent steps, sensors are only allowed to be positioned on the ceiling pointed down, centered at a grid square.  That is, if the grid squares are labeled from 1 to $n$, our optimization variables $x_1, \dots, x_n$ are binary, where $x_i = 1$ means a sensor is placed at square $i$, and is 0 otherwise.  To ensure that quantization errors do not affect counting performance, we use the rule of thumb that the edge length of a grid square should be no larger than 1/5 of the edge length of the sensor field of view at the floor.  For example, for a sensor with 60$^\circ$ angle and a ceiling height of 2.5m, the field of view width at the floor is 2.1m and the maximum grid square width is 42cm.

Each grid square inherits the label with the majority area inside it.  Of critical importance are grid squares containing zone boundaries; we define a binary vector $\mathbf{b} \in \bee{n}$ where $b_i = 1$ if grid square $i$ is on a zone boundary and 0 otherwise.

\subsection{Generating Occupant Trajectories}

In this section, we simulate a large number of realistic paths between pairs of areas of interest. After floor plan labeling, each area of interest contains a set of grid points. Similar to \cite{vlasenko_smart-condo_2015}, for each pair of areas of interest, we choose one point from each set, and treat them as the starting and ending points of a path through walkable space.  A shortest-path algorithm such as the A* algorithm \cite{klein2003parsing} will always find exactly the same path for a given pair of endpoints; however, humans will naturally have some variation in routes between the same two points.  Thus, we incorporate randomness and penalties as described below to introduce realistic variations in the paths.

\begin{figure}[htbp!]
    \centering
    \begin{tabular}{cc}
        \includegraphics[width = 4.5in]{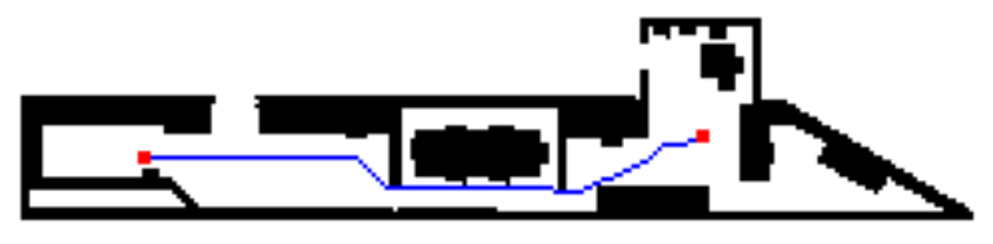} \\ 
        (a)\\
        \includegraphics[width = 4.5in]{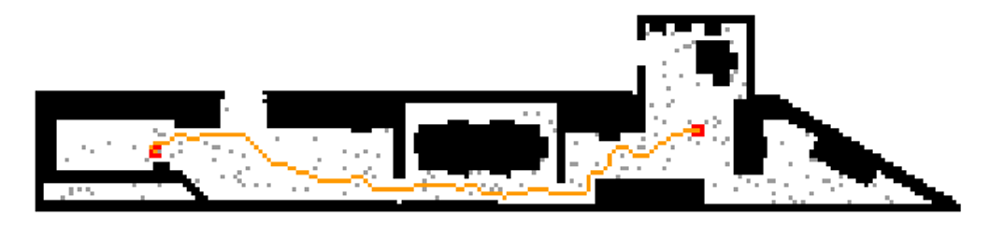} \\
        (b) \\
        \end{tabular}
    \caption{Shortest paths between the same two endpoints in cases that without random obstacles (a) and with random obstacles (b).}
    \label{fig:obs_path}
\end{figure}

First, before generating a path between a pair of endpoints, we randomly block a fraction of grid squares to create obstacles.  This results in a random ``detour'' for each run, introducing variations.  Based on our observations and comparisons with real behavior, blocking 10\% of the walkable grid squares gives good results. Figure~\ref{fig:obs_path} illustrates an example showing how random obstacles (grey blocks) affect the shortest paths between the same endpoints. 



\begin{figure}[htbp!]
    \centering
    \includegraphics[width=\textwidth]{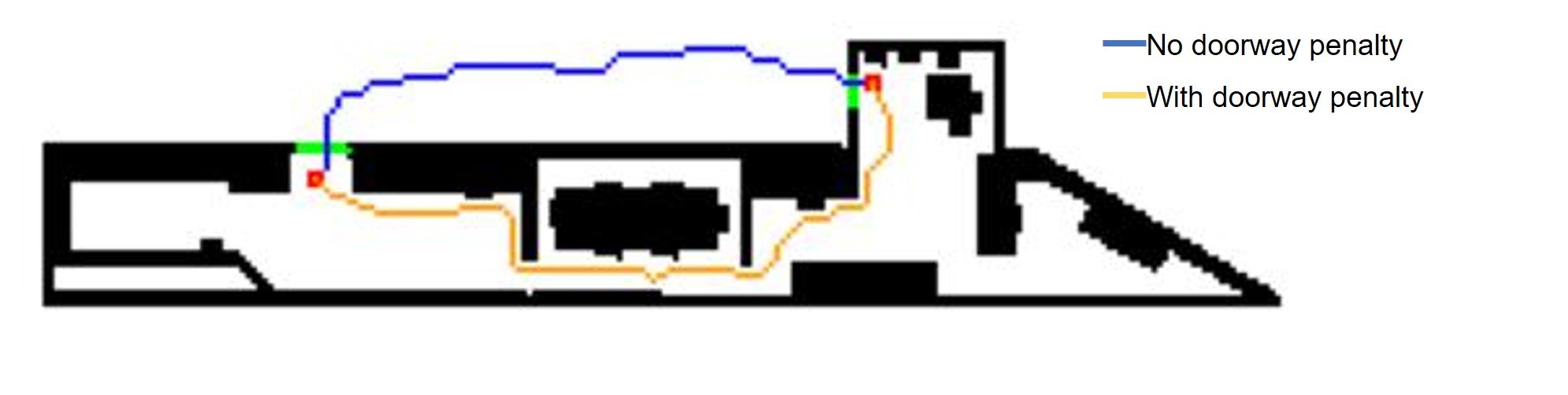}
    \caption{Shortest paths between the same two endpoints without the doorway penalty (blue) and with the doorway penalty (yellow). }
    \label{fig:f3_path_door}
\end{figure}

Second, even though the shortest path between two endpoints may involve exiting and re-entering the office suite, real people are unlikely to take this route. We thus add a penalty for passing through an exterior door equivalent to walking an extra $a$ meters (in our case, $a=3$).  Figure \ref{fig:f3_path_door} illustrates this situation. 

\begin{figure}[htbp!]
    \centering
    \includegraphics[width=\textwidth]{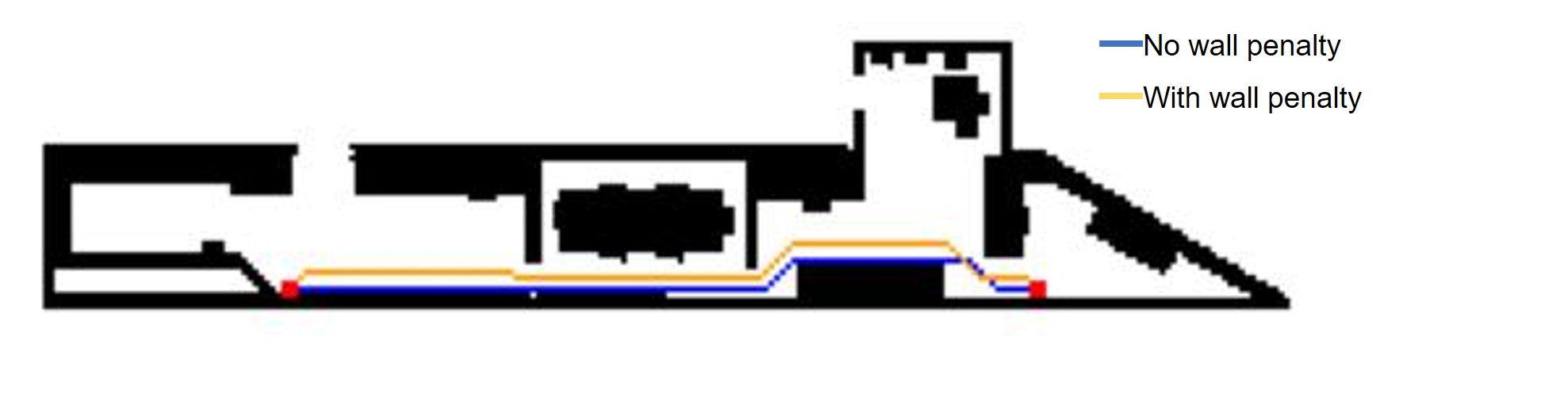}
    \caption{Shortest paths between the same two endpoints without the wall penalty (blue) and with the wall penalty (yellow).} 
    \label{fig:f4_path_way}
\end{figure}

Third, we observed that mathematically shortest paths will often hug the edges of openings and walls, instead of going through gaps with space on either side.  To more accurately model how humans walk, we add a wall penalty such that walking within $b$ cm of a wall is equivalent to $k$ times the actual distance (in our case, $b = 50$ and $k = 1.2$).  Figure \ref{fig:f4_path_way} illustrates this modification. 

\begin{figure}[htbp!]
    \centering
    \includegraphics[width=\textwidth]{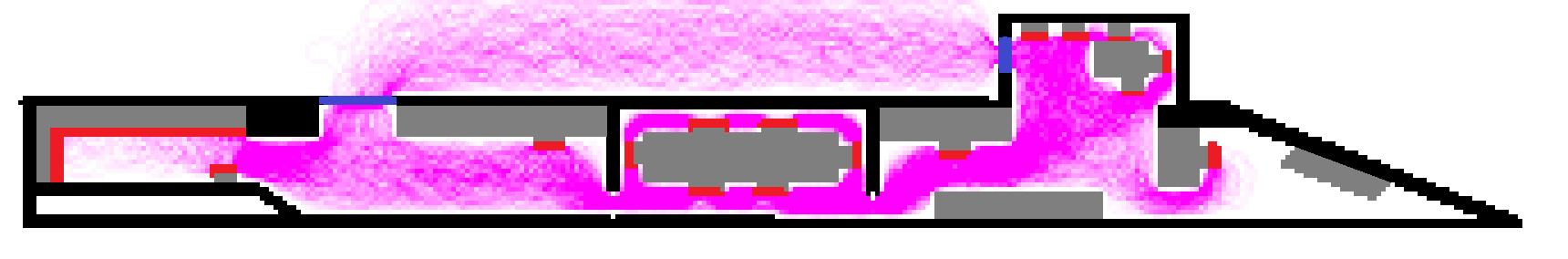}
    \caption{A heatmap of 3000 randomly generated occupant trajectories in an office suite.}
    \label{fig:f2_mobility_pattern}
\end{figure}

Given these considerations, we now compute a large number of trajectories between random pairs of areas of interest.  The number of paths should be proportional to the area of the environment and the complexity of the zone boundaries. In our running example of a $120 m^2$ office suite, we simulated  3000 paths, choosing each starting and ending point randomly from the possible areas of interest and generating a random selection of obstacles for each run. Figure \ref{fig:f2_mobility_pattern} illustrates a ``heatmap'' of the occupant trajectories estimated in this way. \rnote{Check Fig 5 and subsequent figures to make sure they consistently use the same color code as in Figure 1 (especially the door/zone colors.}

While in our experiments, all areas of interest are equally weighted, we note that this process could be made more complex or realistic.  For example, certain areas of interest could be weighted more heavily (e.g., the restroom is visited more frequently than the kitchen), or occupants could be assigned individual desks and daily agendas.  We will investigate the implications of this assumption (i.e., the actual occupant activity differs from the input model) in Section \ref{sec:results}-G \rnote{double check subsection}.

\subsection{Sensor System Modeling} \label{section3.1}

In our earlier work \cite{lu2021zone}, we designed a sensor system based on the VL53L5 time-of-flight sensor recently announced by ST Microelectronics. We mounted each sensor pod on the ceiling pointed directly downwards. The sensor has a detection range of approximately 5cm to 400cm and a diagonal angle of 60 degrees.   In that paper, we manually placed the sensors at the boundaries of each HVAC zone to make sure that every person that walks through a boundary will be detected by a sensor. We used point clustering to distinguish multiple people and a random forest classifier to estimate the location of each person. Based on experiments in both a physical space and extensive simulations, we measured the error rate of our counting system to be around 0.4\%.

We will describe our results in terms of this system for the rest of the paper.  However, we emphasize that the sensor placement algorithm we propose here is not tied to this specific system.  As long as the field of view of each sensor is known, the algorithm described here can be applied.

For a given sensor system, we now create a binary matrix $G \in \bee{n \times n}$ such that $G(i,j) = 1$ if grid square $i$ on the floor is always visible to a sensor placed at grid square $j$ on the ceiling (and 0 otherwise).  We note that this coverage can be affected by furniture or doors (e.g., if a door is open it may prevent a region of floor from being directly observed).

\section{Sensor Placement Optimization}
\label{sec:opt}
In this section, we formally define our optimization problem.  The objective is to place sensors to maximally cover occupant trajectories that are related to zone transitions.  That is, we are not concerned with sensing occupant motion located entirely within a single zone since this has no impact on HVAC control.  

\begin{figure}[htbp!]
    \centering
    \includegraphics[width=\textwidth]{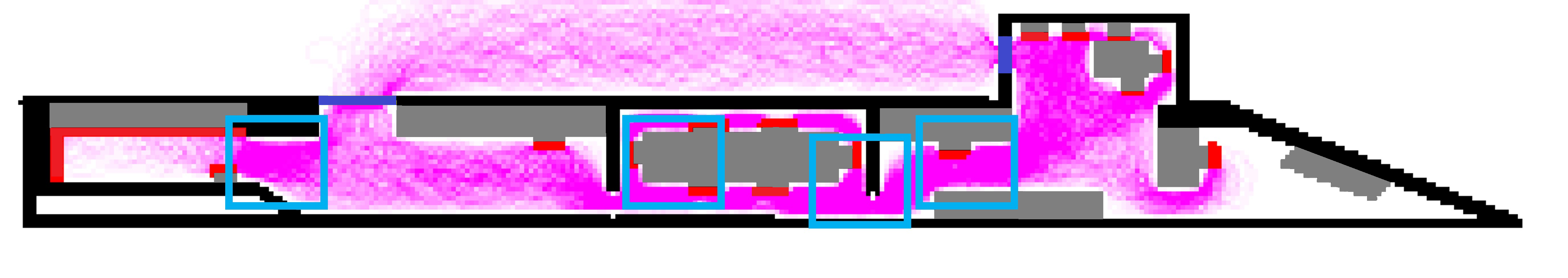}
    \caption{Sensor placement (blue boxes) based on the strategy of covering the densest cells.}
    \label{fig:figure6}
\end{figure}

\begin{figure}[htbp!]
    \centering
    \includegraphics[width=\textwidth]{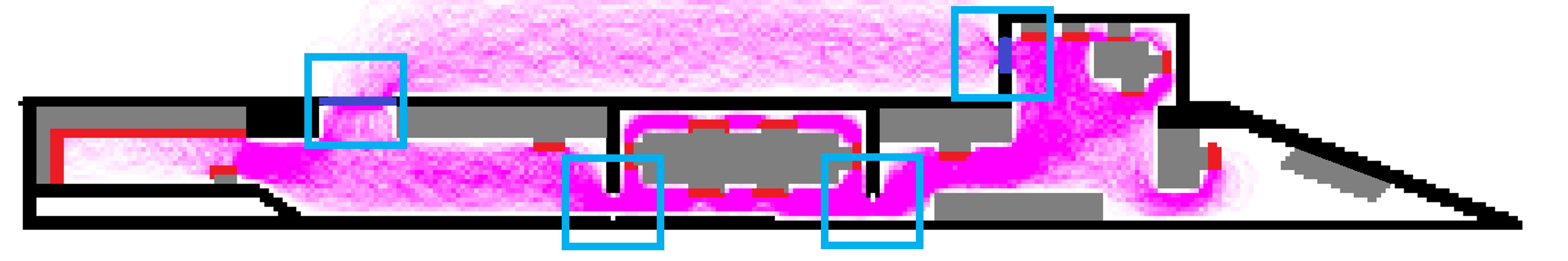}
    \caption{Sensor placement (blue boxes) based on the strategy of maximally covering the intersection points between occupant trajectories and doorways.}
    \label{fig:figure7}
\end{figure}

One of the common approaches to indoor OSP problems is to maximize the critical area coverage \cite{wu_sensor_2020,fanti_smart_2017}. In our context, this would mean that our sensors should cover the area with the densest paths, illustrated in Figure \ref{fig:figure6}.  However, this placement strategy is sensitive to the motion located within a single zone and hence, decreases the coverage rate for trajectories near the doorways. Another approach is that we only consider the intersection points between occupant trajectories and doorways, illustrated in Figure \ref{fig:figure7}. However, this approach will only install sensors directly centered over zone boundaries, which only works well when the number of sensors is no less than the number of boundaries.

\begin{figure}
\begin{tabular}{cc}
\includegraphics[width = 2.5in]{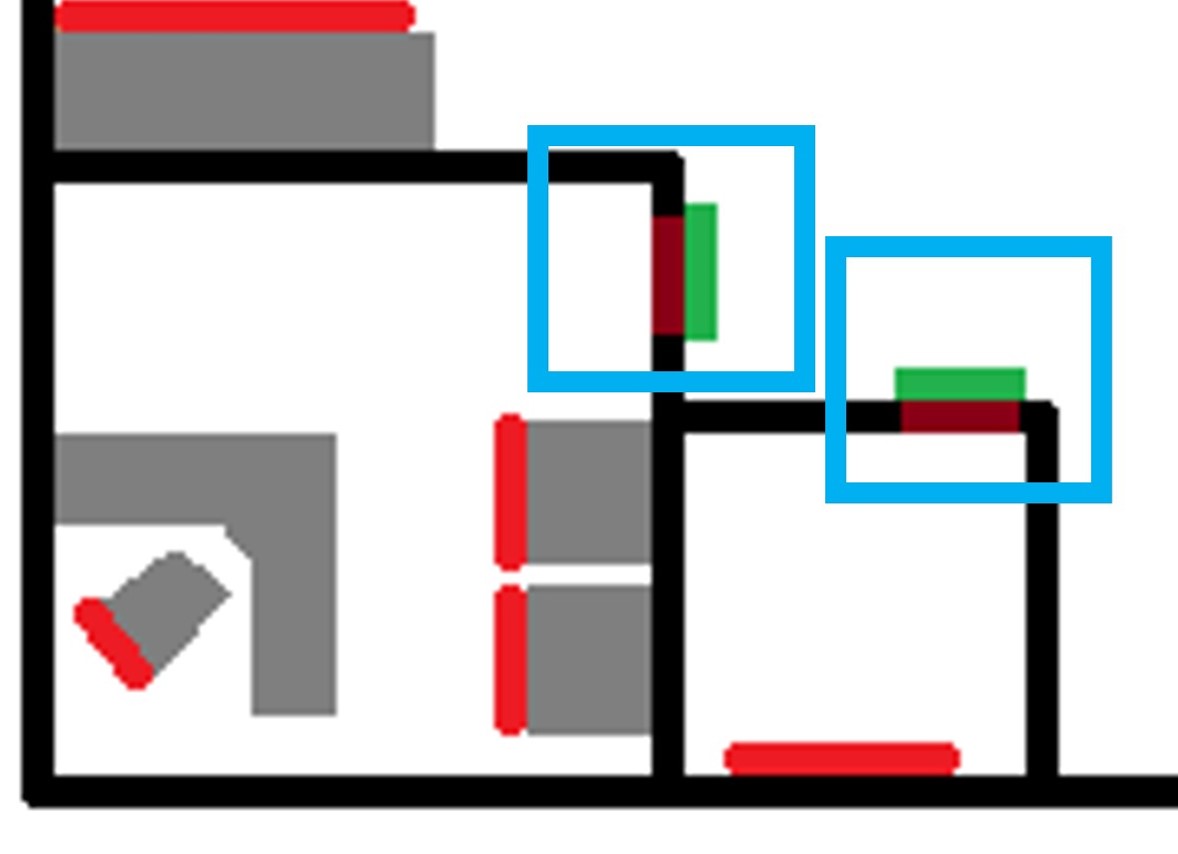} & 
\includegraphics[width = 2.5in]{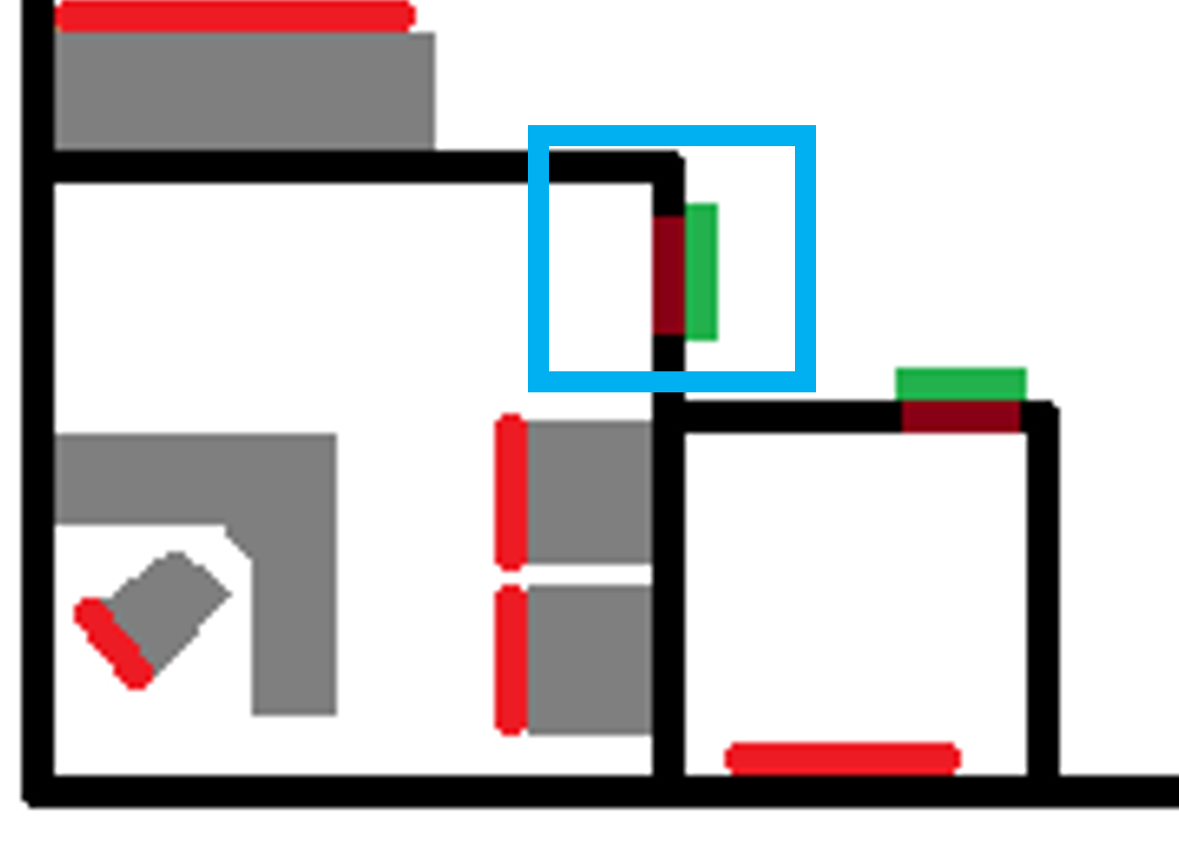} \\
(a) & (b) \\
\includegraphics[width = 2.5in]{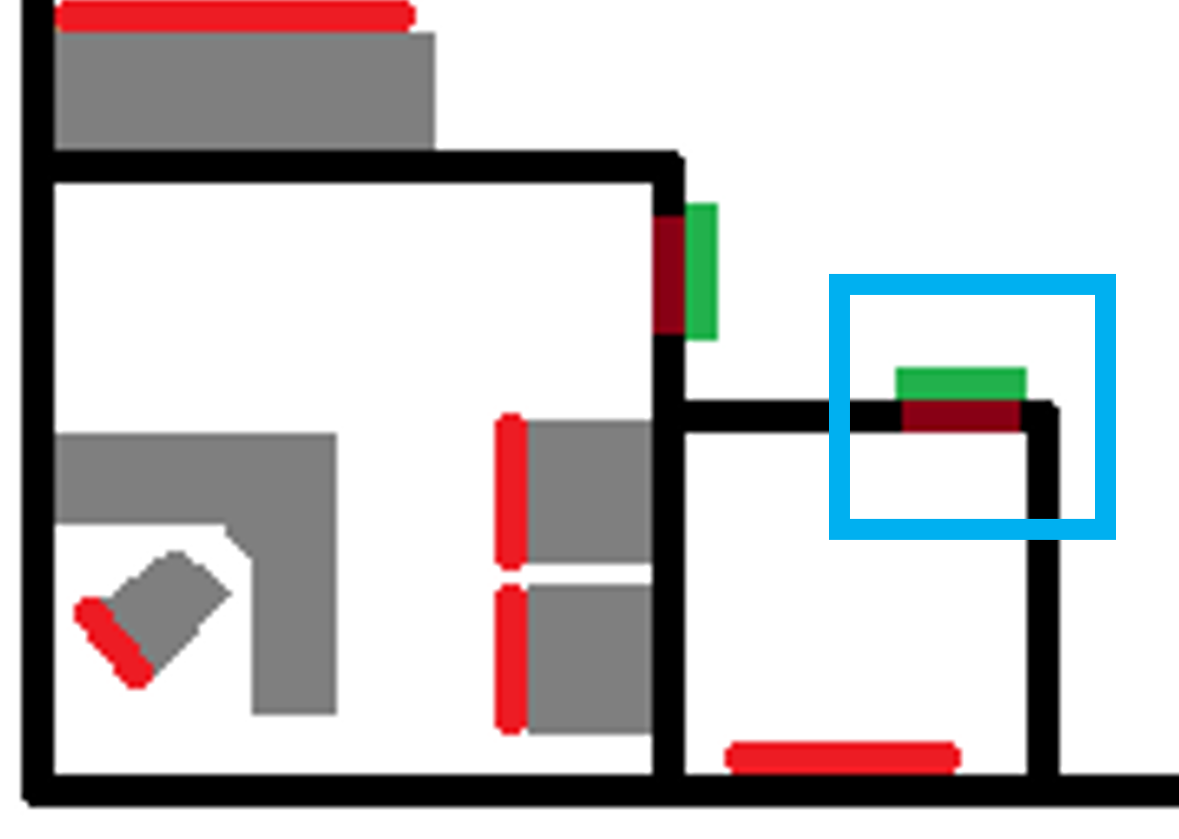} & 
\includegraphics[width = 2.5in]{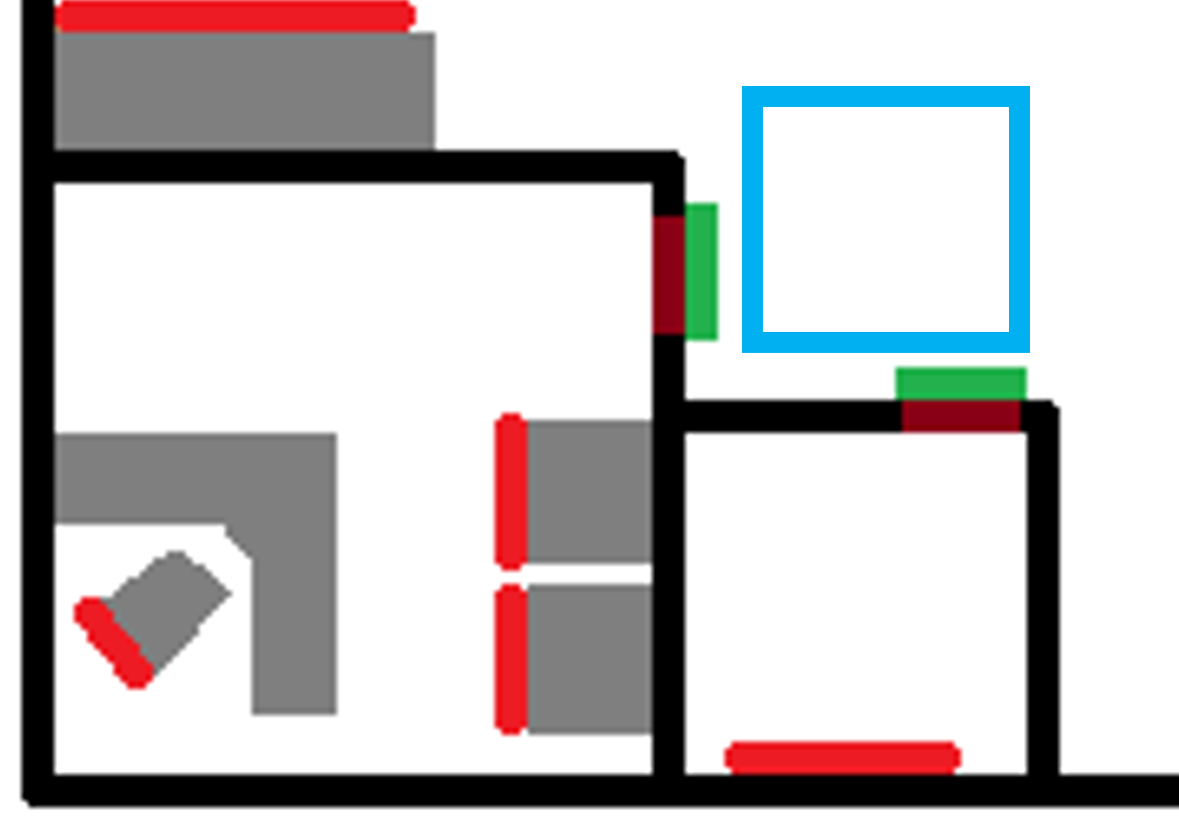} \\
(c) & (d)
\end{tabular}
\caption{Shortcomings of the strategy of maximally covering the intersection points between trajectories and doorways, in a scenario with two zone boundaries. (a) Two sensors; both zones are independently covered.  (b) and (c)  One sensor; one zone is not sensed.  (d) A better placement strategy with one sensor to cover both zones.  \rnote{Note that I put all the caption info into the bottom caption; I prefer this and it's also easier syntax.}}
\label{fig:figure8}
\end{figure}

Figure \ref{fig:figure8} illustrates the problem with this approach. When the sensor number is equal to the number of doorways (shown in Figure \ref{fig:figure8}a), this approach can cover all the critical areas. However, when the number of sensors is less than the number of doorways (shown in Figures \ref{fig:figure8}b and \ref{fig:figure8}c), at least one entry cannot be covered. Our key observation is that if we also consider the trajectories near the doorway and position the sensor as shown in Figure \ref{fig:figure8}d, we can determine the zone transitions for both entries at once. We now describe how to encourage this behavior mathematically.

\begin{figure}
\begin{tabular}{cc}
\includegraphics[width = 3in]{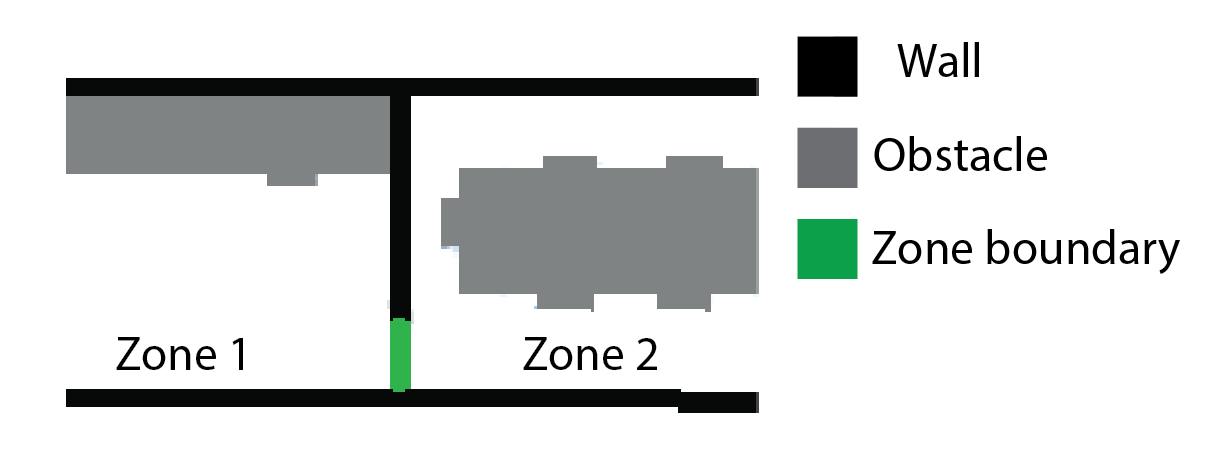} &
\includegraphics[width = 3in]{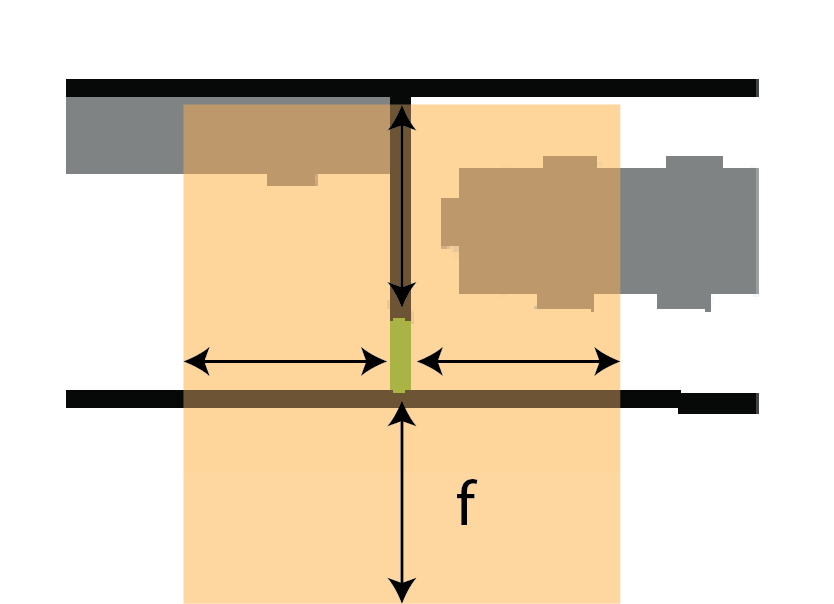} \\
(a) & (b) \\
\includegraphics[width = 3in]{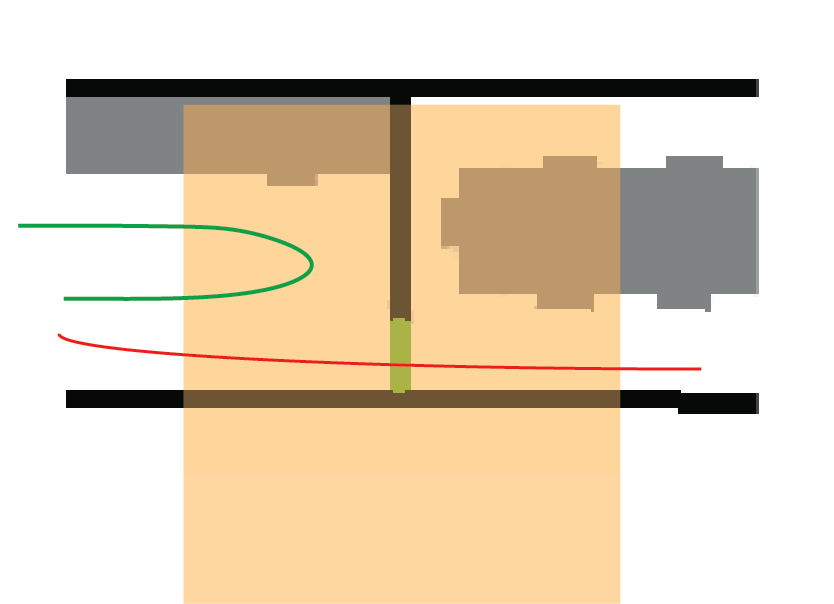} &
\includegraphics[width = 3in]{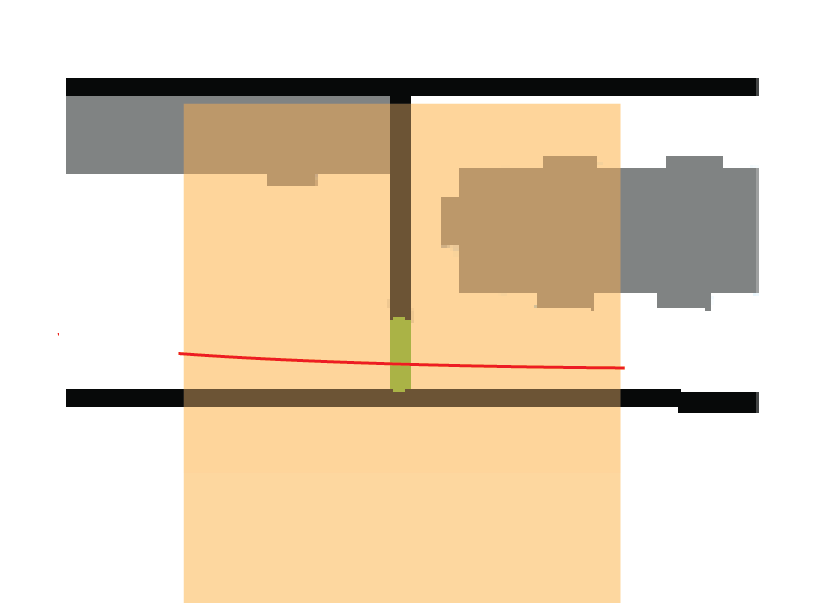} \\
(c) & (d)
\end{tabular}
\caption{Zone boundary dilation and path filtering.  (a) Example layout with one zone boundary.  (b) The zone boundary is dilated by $f$ meters.  (c) Paths intersecting the dilated zone boundary.  (d) Only path segments that transition the zone boundary are retained.}
\label{fig:path_select}
\end{figure}

As illustrated in Figure \ref{fig:path_select}, the basic idea is to dilate the zone boundaries by $f$ m, and keep only those trajectory segments that (a) intersect the dilated region and (b) pass through a zone boundary.  In this way, we preserve only the simulated path segments germane to zone transitions that could be detected by a nearby sensor. When $f$ approaches infinity, this approach is equivalent to the densest path coverage strategy  (Figure \ref{fig:figure6}) and when $f=0$, this approach degenerates to the maximum intersection points method (Figure \ref{fig:figure7}). According to experimental experience, we select $f$ equal to the edge length of the sensor FOV.  \rnote{Fig 9 edits: center text in (a).  remove red and green line captions in (c).}


At the end of this process, we have a total of $p$ partial trajectories, and we define a binary matrix $P \in \bee{n \times p}$ such that if path segment $j$ passes through grid square $i$, $P(i,j) = 1$, and is 0 otherwise.

We can now define our optimization problem in terms of the vectors and matrices previously introduced in Section \ref{sec:traj}: 

\begin{itemize}
    \item $\mathbf{x \in \bee{n}}$, the binary vector to be determined that corresponds to the final sensor positions;
    \item $\mathbf{b \in \bee{n}}$, the binary vector of boundary locations;
    \item $\mathbf{G \in \bee{n \times n}}$, the binary matrix defining sensor coverage;
    \item $\mathbf{P \in \bee{n \times p}}$, the binary matrix defining occupant trajectories.
\end{itemize}

The sensor placement optimization problem is then:

\begin{align}
    \max_{\mathbf{x},\mathbf{y}, \mathbf{z}} & \sum_{j=1}^p z_j \label{eq_ILP_obj} \\ 
    s.t. & \sum_{i=1}^n x_i \leq k \label{eq_ILP_2} \\ 
    & \mathbf{y} \leq \mathbf{Gx} \label{eq_ILP_3} \\ 
    & \mathbf{z} \leq \mathbf{Py} \label{eq_ILP_4} \\ 
    & \mathbf{x} \in \bee{n}, \mathbf{y} \in \bee{n}, \mathbf{z} \in \bee{p} \label{eq_ILP_5}
\end{align} 

\begin{figure}[htbp!]
    \centering
    \includegraphics[width=\textwidth]{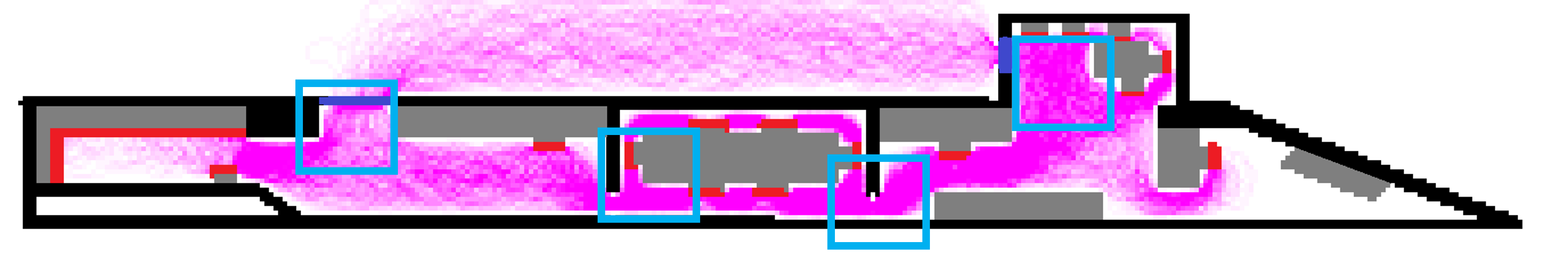}
    \caption{Sensor placement (blue boxes) with our proposed method.}
    \label{fig:figure10}
\end{figure}

This problem involves binary auxiliary variables $\mathbf{y} \in \bee{n}$ and $\mathbf{z} \in \bee{p}$.  Working through the constraints, \rf{eq_ILP_2} ensures that the total number of sensors is no larger than $k$, defined by the user. In \rf{eq_ILP_3}, $\mathbf{Gx}$ is the total number of times each grid square is covered by a candidate assignment, forcing $y_i$ to be 1 if grid square $i$ is covered and 0 otherwise.  Then in \rf{eq_ILP_4}, $\mathbf{Py}$ is the total number of times each trajectory is covered by a candidate assignment, forcing $z_j$ to be 1 if trajectory $j$ is covered and 0 otherwise.  Finally, the objective function \rf{eq_ILP_obj} is simply the number of trajectories covered by an assignment, which we seek to maximize. Figure \ref{fig:figure10} illustrates the sensor placement using our proposed approach. We can see that the sensors are not squarely centered over the doorways, while still being in good positions to observed the zone transitions.

The optimization problem defined by \rf{eq_ILP_obj}--\rf{eq_ILP_5} is an integer linear program, which can be solved with the branch and bound method \cite{lawler1966branch}.  In our problems, $n$ ranges from 10000-50000 and $p$ ranges from 1000-3000.  We chose the CBC solver\footnote{\url{https://https://github.com/coin-or/Cbc}} running in Python 3.8, on Intel(R) Xeon(R) E5-2620 v3, 2.40GHz CPU. The typical solution time is around 15 mins, which while not interactive, is also not overly burdensome (e.g., a designer could run many scenarios overnight).

\section{Experimental Results}
\label{sec:results}
In this section, we discuss experimental results to evaluate the performance of our sensor placement method, leveraging a Unity simulation described below.  We also report the impact of differences between the estimated occupant trajectories as opposed to the actual occupant trajectories on the counting performance.  

\subsection{Test Environments}
\begin{figure}[htbp!]
\begin{tabular}{cc}
\includegraphics[width = 2.5in]{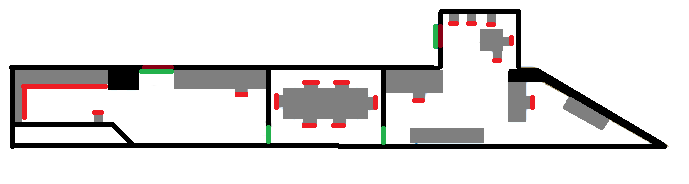} &
\includegraphics[width = 2.5in]{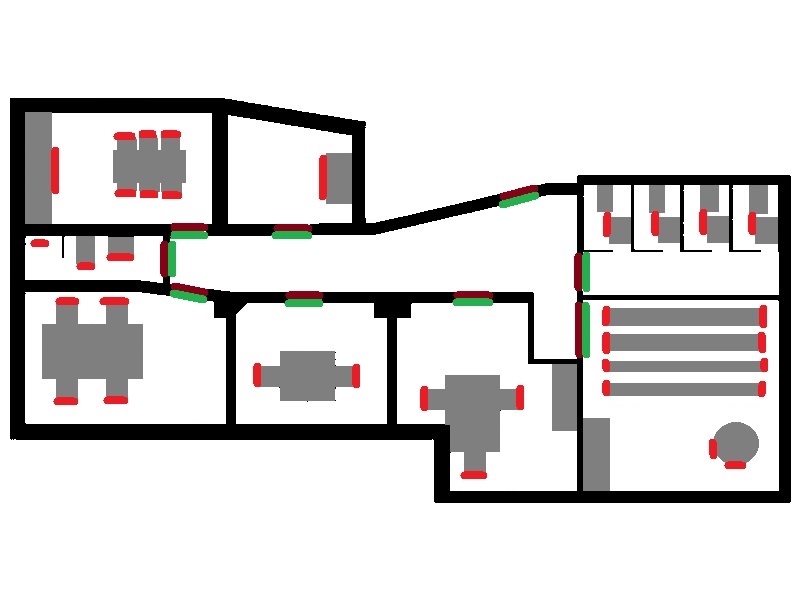} \\
(a) & (b) \\
\includegraphics[width = 2.5in]{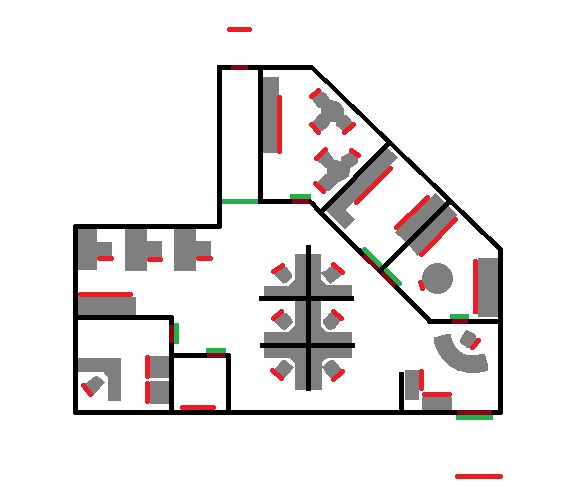} &
\includegraphics[width = 2.5in]{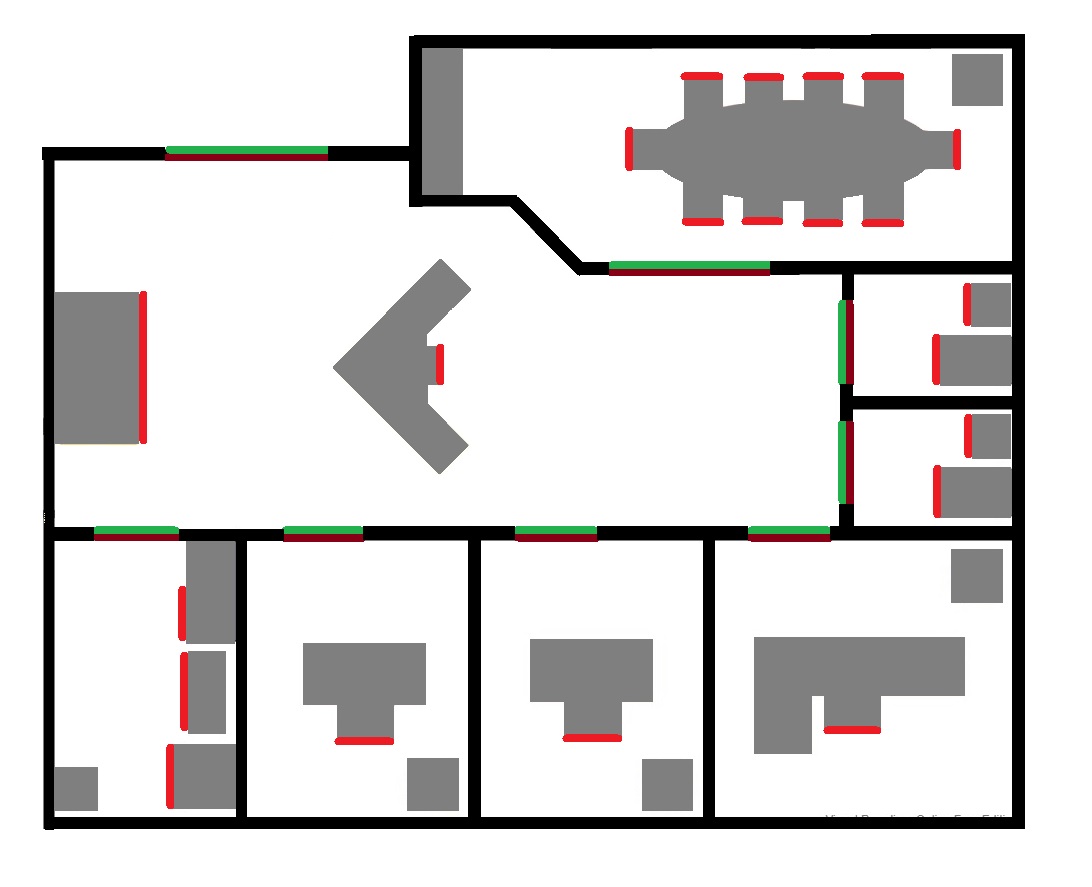} \\
(c) & (d)
\end{tabular}
\caption{Several of the floor plans used to test the proposed method.}
\label{fig:figure11}
\end{figure}

To evaluate the proposed OSP method, we built 6 different office environments with different scales and layouts. We consider a diverse set of offices to test the generalization ability of our method. For example, one office floor plan has a long, narrow corridor and zone boundaries that are far away from each other (Figure \ref{fig:figure11}a). Another office also has a long and narrow corridor but with zone boundaries close together (Figure \ref{fig:figure11}b). We also built office environments with large middle spaces and irregular (Figure \ref{fig:figure11}c) and regular (Figure \ref{fig:figure11}d) zone boundaries.  The evaluation results for all these environments will shown in the rest of this section.

\subsection{Unity 3D Simulation}
\begin{figure}[htbp!]
    \centering
    \includegraphics[width=.7\textwidth]{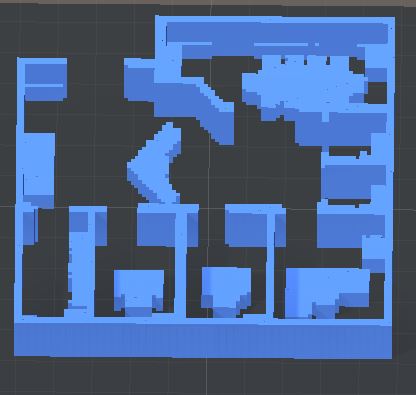}
    \caption{Unity 3D simulation automatically created from the floor plan in Figure \ref{fig:figure11}d.}
    \label{fig:unity}
\end{figure}

To evaluate the counting performance of a given sensor placement, we use the Unity 3D game engine to simulate the office environment and the occupants' behaviors.  We devised a program that automatically produces a Unity simulation of an office space from an image of its floor plan (e.g., Figure \ref{fig:figure11}d) by extruding walls and furniture to create a 3D environment, as illustrated in Figure \ref{fig:unity}.

In addition to the space itself, we also automatically generate realistic simulations of occupant behavior in the spaces.  As described in \cite{lu2021zone}, we generate animated 3D human workers that can walk, sit, and open doors in the space as they travel between randomly chosen areas of interest, which can be the same as we used in Section \ref{section3.2}, or different to test the impact of different areas of interest on counting performance.

As described in \cite{lu2021zone}, we created a digital twin of our time-of-flight sensing system that can be directly applied to evaluate the zone counting performance for a given sensor placement.  Even though the actual counting algorithm is not part of the optimization problem \rf{eq_ILP_obj}--\rf{eq_ILP_5}, our goal is to show that counting performance correlates well with the objective of trajectory coverage.

\subsection{Evaluation Metric}

Our overall metric for performance is the windowed classification rate \cite{cokbas2020low} defined by 

\begin{equation} \label{eq7}
    CCR = \frac{TP}{TP + FP + FN}
\end{equation}

Here, $TP$ is the number of true positive detections, $FP$ is the number of false positive detections, and $FN$ is the number of false negatives or misses.  A detection is considered to be correct if it occurs within $w$ seconds of an actual zone transition. In this work, we set $w = 2s$ (30 ToF frames). This value is based on the reasoning that if the walking speed of a person is $v = 1m/s$, it takes about $f/v = 2$ seconds to walk from the edge of the dilated boundary to the middle point of the boundary.



\subsection{Results}

\begin{figure}[htbp!]
    \centering
    \includegraphics[width=.7\textwidth]{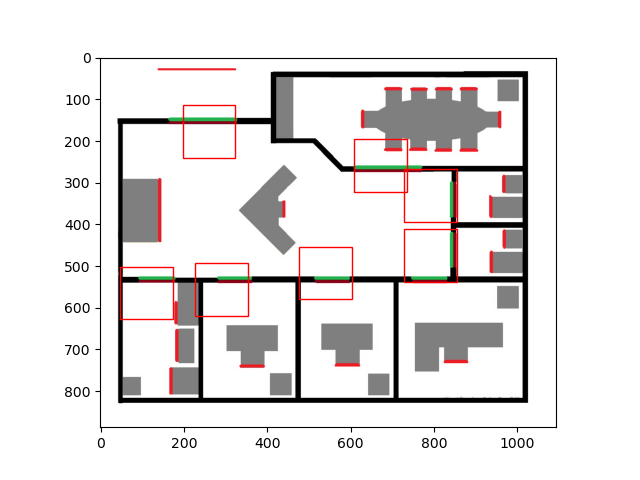}
    \caption{Sensor placement using proposed method with 7 sensor in Office 4.}
    \label{fig:fig14}
\end{figure}

Figure \ref{fig:fig14} illustrates a typical result for our approach in the Office 4 environment (Figure \ref{fig:figure11}d) with 7 sensors. 

The critical question is how the actual performance for a given sensor placement (based on the measured CCR from the Unity simulation) correlates with the value of the objective function (\ref{eq_ILP_obj}) at convergence.  
Table \ref{tab:table1} summarizes these results. For each office environment, we illustrate the floor plan with labeled interest points, and a graph of our objective function value for a given number of sensors (red line) compared with the measured CCR performance of the corresponding sensor placement (blue line).  We can observe that there is good agreement between the two curves in each environment, especially as the number of sensors increases. in each graph, label red line as ``objective function value''.

\begin{table}[htbp!]
    \centering
    \begin{tabular*}{.99\textwidth}{| c | c | c |}
        \hline \centering
        room size & floor plan & Coverage rate and CCR \\
        \hline \centering
        \makecell[c]{Office 1 \\ $4m \times 21m$}&
        \begin{minipage}{.4\textwidth}\centering \includegraphics[width=.9\linewidth, height=.5\linewidth]{pictures/fp1_code.png} \end{minipage} & \begin{minipage}{.4\textwidth} \includegraphics[width=.9\linewidth, height=.5\linewidth]{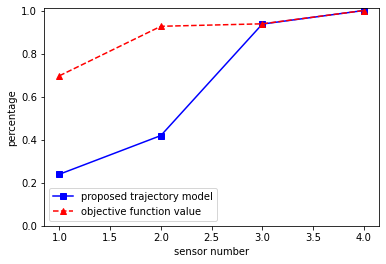} \end{minipage}                                            \\
        \hline \centering
        \makecell[c]{Office 2 \\ $20m \times 20m$}&
        \begin{minipage}{.4\textwidth} \includegraphics[width=.9\linewidth, height=.5\linewidth]{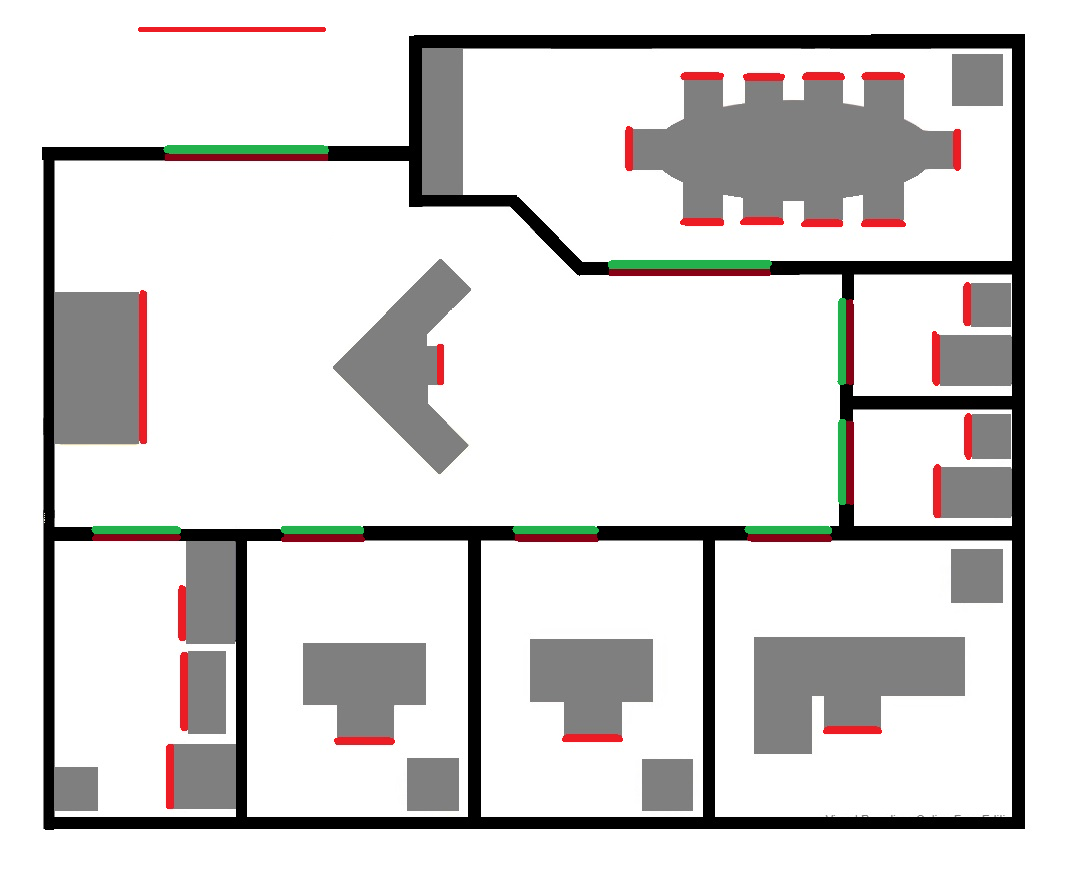} \end{minipage} & \begin{minipage}{.4\textwidth} \includegraphics[width=.9\linewidth, height=.5\linewidth]{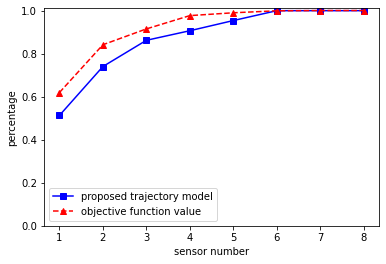} \end{minipage}                                                  \\
        \hline \centering
        \makecell[c]{Office 3 \\$25m \times 40m$}&
        \begin{minipage}{.4\textwidth} \includegraphics[width=.9\linewidth, height=.5\linewidth]{pictures/fp3_code.jpg} \end{minipage} & \begin{minipage}{.4\textwidth} \includegraphics[width=.9\linewidth, height=.5\linewidth]{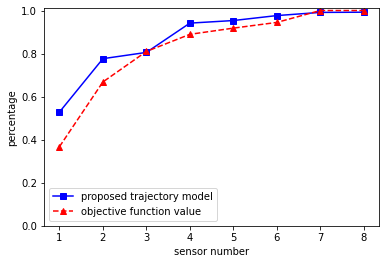} \end{minipage}                                                \\
        \hline \centering
        \makecell[c]{Office 4 \\$20m \times 20m$}&
        \begin{minipage}{.4\textwidth} \includegraphics[width=.9\linewidth, height=.5\linewidth]{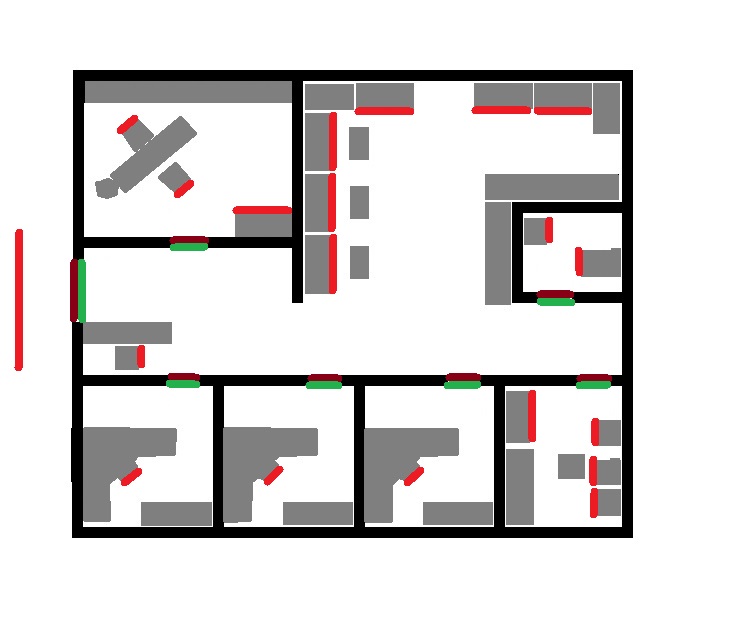} \end{minipage} & \begin{minipage}{.4\textwidth} \includegraphics[width=.9\linewidth, height=.5\linewidth]{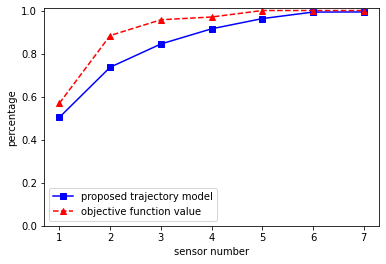} \end{minipage}                                                \\
        \hline \centering
        \makecell[c]{Office 5 \\$14m \times 10m$}&
        \begin{minipage}{.4\textwidth} \includegraphics[width=.9\linewidth, height=.5\linewidth]{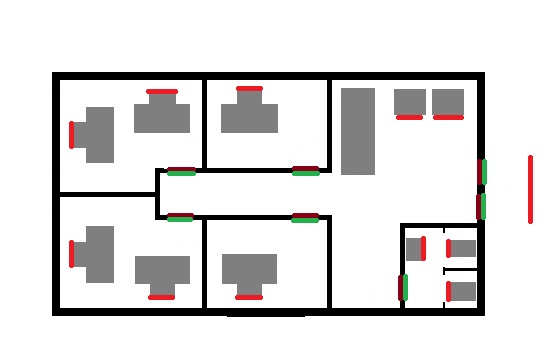} \end{minipage} & \begin{minipage}{.4\textwidth} \includegraphics[width=.9\linewidth, height=.5\linewidth]{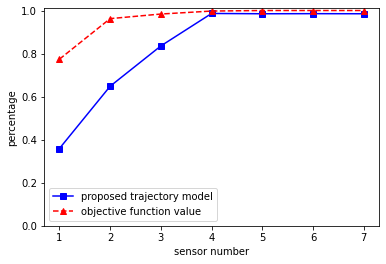} \end{minipage}                                                \\
        \hline \centering
        \makecell[c]{Office 6 \\$15m \times 19m$}&
        \begin{minipage}{.4\textwidth} \includegraphics[width=.9\linewidth, height=.5\linewidth]{pictures/fp6_code.png} \end{minipage} & \begin{minipage}{.4\textwidth} \includegraphics[width=.9\linewidth, height=.5\linewidth]{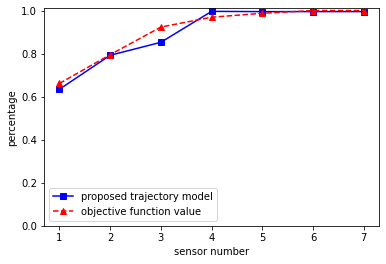} \end{minipage}    \\      \hline
    \end{tabular*}
    \caption{Coverage rate and CCR for different layout of office environments.}
    \label{tab:table1}
\end{table}



\subsection{Role of the Dilation Parameter}
As we mention in section~\ref{sec:opt}, to eliminate the effects of extraneous trajectories without confining the sensor near the door, we introduce the dilation parameter $f$ and recommend it set as the same of the edge length of FOV of sensors. In this subsection we will discuss about the effect of $f$ on the accuracy rate.

\begin{figure}[htbp!]
    \centering
    \includegraphics[width=.7\textwidth]{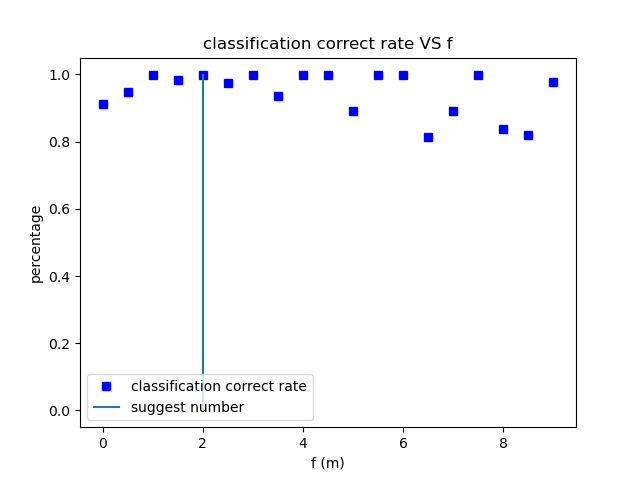}
    \caption{The variance of classification correct rate against parameter f (m).}
    \label{fig:fig15}
\end{figure}

We select office 4, illustrated as Figure~\ref{fig:figure11}d. To reflect the phenomenon mentioned in Figure~\ref{fig:figure8}, we set the sensor number is 6, two less than the number of zone boundaries. The classification correct rate (CCR) against $f$ is illustrated as Figure~\ref{fig:fig15}. We can learn that, the CCR increasing as $f$ increasing when $f<1m$. When $1m<f<3m$, the CCR oscillates at higher values. But when $f>3m$, due to the effects of extraneous trajectories, the accuracy rate starts to fluctuate wildly. This fits the explanation we give in section~\ref{sec:opt} well.

\subsection{Choosing the Number of Sensors}
\begin{figure}[h!]
\subfloat[Limited budget ($\alpha$ = 0.05) \label{fig:fig15a}]{\includegraphics[width = 2.5in]{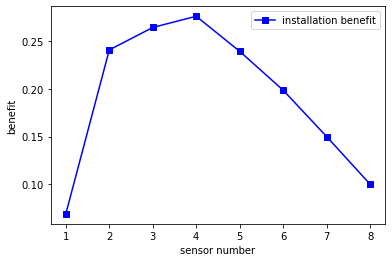}} 
\hfill
\subfloat[High accuracy requirement ($\alpha$ = 0.005) \label{fig:fig15b}]{\includegraphics[width = 2.5in]{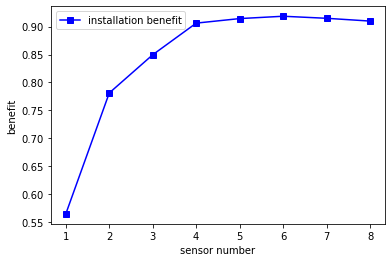}}
\caption{Choosing the number of sensors in Office 2 by trading off performance and cost.}
\label{fig:figure15}
\end{figure}

A natural question is how to choose the number of sensors to balance the counting accuracy and installation cost.  As demonstrated in the previous section, the objective function value (and related real-world performance) in a given environment monotonically increases the more sensors we are allowed. In practice, the installation cost for a sensor network (including both materials and labor) is basically linear in the number of sensors, so it makes sense to choose the number of sensors $n$ by maximizing a benefit function computed as the objective function (\ref{eq_ILP_obj}) minus $\alpha n$, i.e.,  a penalty proportional to $n$.  A small value of $\alpha$ corresponds to a high-budget, high-accuracy requirement, while a large value of $\alpha$ trades off performance for reduced cost.  Using Office 2 as an example, Figure \ref{fig:figure15} illustrates that if we were to set $\alpha = 0.05$ we would choose 4 sensors, while if $\alpha = 0.005$, we would choose 6 sensors.  

\subsection{Effects of Differing Trajectory Models}
Finally, we discuss the situation when the modeled occupant trajectories differ from the actual occupant trajectories.  Since our proposed approach is based on simulations of occupant motion, it's important to understand what happens if the simulations are not totally accurate.

\begin{figure}[htbp!]
    \centering
    \subfloat[Office 2\label{fig:fig16a}]{\includegraphics[width = .45\textwidth]{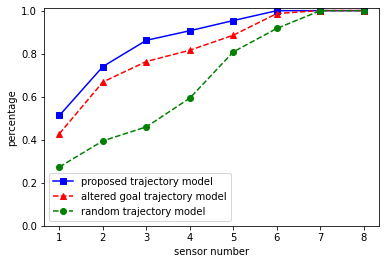}} 
    \subfloat[Office 6\label{fig:fig16b}]{\includegraphics[width = .45\textwidth]{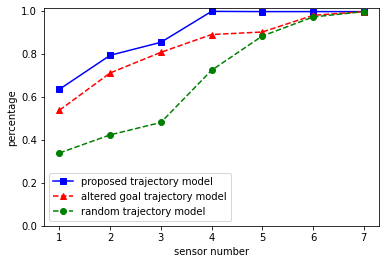}}
    \caption{Comparing CCR performance when the users move exactly according to the input model (blue lines), to different areas of interest (red lines), or totally random locations (green lines).}
    \label{fig:fig17}
\end{figure}

We consider Office 2 and Office 6, where based on Table I the match between objective function value and performance is the best, to study this issue. First, we selected a new and different set of areas of interest in each office, and let the simulated occupants move between them as described in Section V-B.  We then evaluate the CCR performance over these new trajectories using the sensor positions optimized by the proposed method for the old trajectories.  These are graphed as red lines in Figure \ref{fig:fig17}.  We also considered an extreme case in which each occupant of the room randomly picks a reachable point in the environment and moves to it. After reaching this point, another point will be randomly selected, and the process will be repeated.  This random behavior, while unrealistic, can be viewed as a lower bound on the performance of the deployed system.  This scenario is graphed as green lines in Figure \ref{fig:fig17}.  As expected, when the occupant trajectories are reasonable and similar to the original model, the difference in performance (blue vs.~red) is low.  The difference in performance is much larger for the unrealistic situation of random trajectories (blue vs.~green) but at higher sensor counts, the performance across all behaviors is high. In the figure legend, swap the order of the red and green lines.

\section{Conclusions}
 \label{sec:conclusions}

Real-time zone occupancy counting is a key prerequisite for energy savings in commercial office environments.  In this paper, we investigated the optimal sensor placement problem for commercial office environments, creating automatic sensor layouts without requiring installer expertise or manual trial-and-error. 
Our approach was based on using a modified $A^*$ algorithm to simulate the trajectories of occupants in the environments, and formulating an ILP problem to maximize the sensor coverage of trajectories near zone transitions.   Our results on a diverse set of commercial office environments shows excellent agreement with the performance that would be obtained from actual sensor deployments, and we provided guidelines for choosing the optimal number of sensors.  

The results in Section \ref{sec:results}-G demonstrate that differences between the model trajectories and the actual ones have a greater impact when the number of sensors is small.  In future work, we plan to investigate methods for long-term human trajectory forecasting to better estimate occupant positions and velocities. We will leverage related research in trajectory forecasting in outdoor environments \cite{mangalam2021goals,loukkal2021driving}. Fault tolerance of the optimized placements is also a important research direction for practical applications. Since dead batteries and sporadic communication dropouts will always be  problematic in wireless sensor networks; we should incorporate this aspect into our placement algorithm.

\section*{Acknowledgements}
The information, data, or work presented herein was funded in part by the Advanced Research Projects Agency-Energy (ARPA-E), U.S. Department of Energy, under Award Number DE-AR0000942. The views and opinions of authors expressed herein do not necessarily state or reflect those of the United States Government or any agency thereof.


\bibliographystyle{IEEEtran}

\bibliography{myrefs} 

\end{document}